\documentclass{article}
\usepackage{spconf,amsmath,graphicx}
\usepackage{color}
\graphicspath{{snapshots/images/}}

\title{NEURAL NETWORK-BASED AUTOMATIC LIVER TUMOR SEGMENTATION WITH RANDOM FOREST-BASED CANDIDATE FILTERING}
\name{
  Grzegorz Chlebus$^{\star}$ \thanks{We gratefully thank Itaru Endo from the Yokohama City University Medical Center, Yokohama, Japan, for providing data for training of our algorithms.} \qquad
  Hans Meine$^{\star \dagger}$ \qquad
  Jan Hendrik Moltz$^{\star}$ \qquad
  Andrea Schenk$^{\star}$
}

\address{
  $^{\star}$ Fraunhofer Institute for Medical Image Computing MEVIS, Bremen, Germany \\
  $^{\dagger}$ Medical Image Computing Group, University of Bremen, Bremen, Germany \\
}

\begin{document}
%
\maketitle
\begin{abstract}
  We present a fully automatic method employing convolutional neural
  networks based on the 2D U-net architecture and random forest
  classifier to solve the automatic liver lesion segmentation problem
  of the ISBI 2017 Liver Tumor Segmentation Challenge (LiTS).  In
  order to constrain the ROI in which the tumors could be located, a
  liver segmentation is performed first.  For the organ segmentation
  an ensemble of convolutional networks is trained to segment a liver
  using a set of 179 liver CT datasets from liver surgery planning.
  Inside of the liver ROI a neural network, trained using 127
  challenge training datasets, identifies tumor candidates, which are
  subsequently filtered with a random forest classifier yielding the
  final tumor segmentation.  The evaluation on the 70 challenge test
  cases resulted in a mean Dice coefficient of 0.65, ranking our
  method in the second place.
\end{abstract}
\section{Introduction}
\label{sec:intro}
Primary and secondary liver lesions can have heterogeneous shape and
appearance in image data and thus are very challenging to be segmented
automatically. Currently, an acceptable quality level can only be
achieved with interactive segmentation approaches.

In this paper, we describe a fully automatic method for liver tumor
segmentation, which employs cascaded CNNs similar to
\cite{Christ2016}, to provide a liver and tumor segmentation.
We additionally filter the tumor candidates by means of shape- and
image-based features computed for each candidate.

The rest of the paper is structured as follows. In
Sec. \ref{sec:liver_segmentation} we describe our method for
extracting a liver mask from a given CT volume, followed by the liver
tumor segmentation outlined in Sec.\,\ref{sec:tumor_segmentation}. In
Sec. \ref{sec:results_discussion}, we report the training time and
inference time per volume of our method together with results achieved
on the challenge test data. We conclude in Sec. \ref{sec:conclusions}.

\section{Automatic Liver Segmentation}
\label{sec:liver_segmentation}

Being able to segment the liver in a CT volume in a robust way plays a
crucial role in our method, as the subsequent tumor segmentation is
constrained to the liver ROI.  For this organ segmentation, we use a
two-step liver segmentation pipeline.  The first step employs an
ensemble of three orthogonal 2D neural networks based on the U-net
architecture~\cite{Ronneberger2015}, the second uses a 3D
U-net~\cite{Çiçek2016} to refine the liver mask.  Both steps are
given in detail in the following.

\subsection{Data and Preprocessing}
We trained the neural networks on a dataset containing 179 liver CT
volumes used for surgery planning. The dataset was split randomly into
two non-overlapping groups comprising 147 and 32 volumes, which were
used for training and validation, respectively. The images were
acquired on different scanners with a resolution of about 0.6\,mm
in-plane and a slice thickness of 0.8\,mm. All cases were annotated by
radiological experts on the venous phase using a slice-wise method
based on live-wire, shape-based interpolation, and interactive contour
correction\,\cite{Schenk2000}.

As preprocessing step, we applied DICOM rescale parameters to get
values in Hounsfield units, and the padding necessary for the
convolutional networks was accordingly done with a fill value of
-1000\,HU.  Resampling was applied only for the first step and will be
described below.

\subsection{Network Architecture and Training}
The 2D U-net model employed in the first step of the liver
segmentation pipeline works on 4 resolution levels, resulting in a
receptive field of 99 voxels. The channel count is 64 on the first
resolution level and is doubled (halved) on each resolution change
along the downscaling (upscaling) path, respectively. All neural
networks described in this work use ReLu activations followed by batch
normalisation and are trained using the Dice loss function and the
Adam optimizer on a GeForce GTX 1080.

For the first step, three instances of the 2D U-net model were trained
using axial, sagittal and coronal slices resampled to isotropic 2\,mm
voxels, respectively.  Training was stopped due to convergence after
about 8~hours each (batch size 10 slices, learning rate 0.005).

The second step had the task of combining the three outputs to a final
mask at full original resolution.  For this, we employed a small 3D
U-net with only two resolution levels and a receptive field of $17^3$
voxels.  This network got four input channels: The softmax output of
the three 2D U-nets resampled to the original resolution and the
original CT volume image, after adapting its value range to a similar
range ($v = \left( v_\text{HU} + 1000 \right) / 1000$).  For this
network we used a batch size of $4\times(42\times42\times42)$ voxels
and a learning rate of 0.00035.  Despite the higher dimensionality and
resolution, convergence with this tiny model was even faster, so
training was again stopped after less than 8 hours.

\subsection{Postprocessing}
We take the biggest connected component after applying a threshold
value of 0.5 to the softmax output of the 3D U-net output to receive
the final liver mask.

\section{Automatic Tumor Segmentation}
\label{sec:tumor_segmentation}
The liver mask serves as input to the two-step tumor segmentation
pipeline to constrain the ROI in which the tumors are searched for.
As discussed in the following sections, the first step utilizes a 2D
U-net to produce an image containing tumor candidates, which are
filtered in the second step using a random forest classifier trained
on shape- and image-based features.

\subsection{Data and Preprocessing}
We used the training data provided by the challenge organizers
amounting to 131 CT volumes from various clinical sites with a
resolution of about 0.8\,mm in-plane and slice thickness of
1.5\,mm. The training data comes with reference liver and tumor
segmentations. For the training of the neural network and random
forest classifier we left 4 flawed cases out (e.g., due to missing
reference tumor segmentation). The remaining 127 cases were divided
into three non-overlapping groups containing 101, 15 and 11 cases,
which were used for training, validation and testing, respectively.

Again, the CT volumes were rescaled to obtain values in Houns\-field units.

\subsection{Network Architecture and Training}
We used the same 2D U-net architecture as in
Sec.\,\ref{sec:liver_segmentation} to train a model for tumor
segmentation. We trained the model with axial slices in the original
resolution with a batch size of 3 and 0.00035 learning rate, which took
about 12\,hours.
The loss was restrained to a slightly dilated liver
mask in order to focus the network on the liver region. Due to the
high class imbalance (tumor class vs. background class), we trained the
network using boundary patches only (i.e., patches containing voxels
of both classes).

\subsection{Postprocessing}
Our tumor segmentation mask is based on the hard classification output
of the neural network, masked with the liver mask. Subsequently, we
compute tumor candidates by extracting 3D connected components of the
result.

The network trained so far does not yet have high specificity,
possibly due to the employed boundary patch sampling
strategy. Therefore, we employ another classifier in order to filter
out false positives among the tumor candidates.  In this step, we
trained a random forest classifier on all tumor candidates produced
for the training cases, with features computed for the raw and a
refined version of the segmentation mask.  The refined version was
created using a stroke-based liver tumor segmentation
algorithm~\cite{Moltz2009} initialised with the biggest diameter of
the raw tumor candidate.  We extracted 46 features for each version of
tumor candidate using object-based image analysis methods based on the
tumor candidate shape, original image intensity statistics and liver
mask distance transformation\,\cite{Schwier2011}. The latter was added
in order to discriminate tumor candidates at the border of the liver
mask.

\section{Results and Discussion}
\label{sec:results_discussion}
We computed tumor segmentations for the 70 challenge test cases,
achieving a mean Dice coefficient of 0.65, which placed our method at
the second place at the LiTS Challenge.  The method needs on average
195\,s for one case (65, 49 and 81\,s for liver segmentation, tumor
segmentation and tumor postprocessing, respectively).

\begin{figure}[t]
  \centering
  \includegraphics[width=8.0cm]{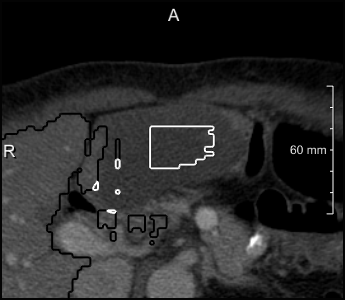}
  \caption{A case where an inaccurate liver mask influenced the tumor segmentation (white line). }
  \label{fig:fig1}
\end{figure}

The masking of tumors with the liver region leads in some cases to
either an insufficient or a completely wrong tumor segmentation
(Fig.\,\ref{fig:fig1}).  We observed, that this problem occurs mainly
for cases where bigger tumors are located at the liver boundary.

\begin{figure}[t]
  \centering
  \includegraphics[width=8.0cm]{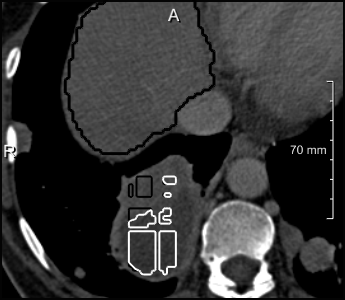}
  \caption{A case where the tumor segmentation included regions outside the liver. }
  \label{fig:fig2}
\end{figure}

The fact that the loss function of the neural network trained to
segment tumors was constrained to the liver mask resulted in a faster
training time. The network did not learn how the voxels outside
the liver mask should be classified, which leads in some cases to
false positives, which cannot be filtered out with our random
forest-based postprocessing (Fig.\,\ref{fig:fig2}).

\begin{figure}[t]
  \centering
  \includegraphics[width=8.0cm]{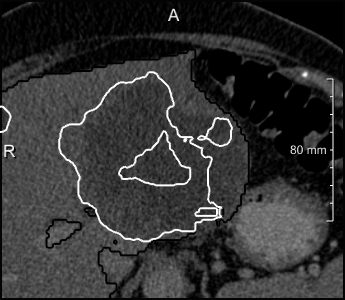}
  \caption{A case where the tumor segmentation network failed to segment the tumor correctly.}
  \label{fig:fig3}
\end{figure}

We also observed that the tumor segmentation network has problems with
segmenting the center voxels of bigger tumors correctly
(Fig.\,\ref{fig:fig3}).  A possible reason for this behavior may be
missing context information thath could no be provided by the the
effective receptive field\,\cite{Luo2016}.

\begin{figure}[!t]
  \centering
  \includegraphics[width=8.0cm]{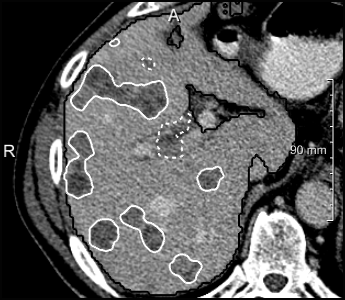}
  \caption{The tumor candidates marked with a dashed white line were classified as false positives.}
  \label{fig:fig4}
\end{figure}

The employed boundary patch sampling strategy results in a higher
sensitivity at a cost of a lower specificity compared to a training on
all patches.  This increases the probability
to find all tumors.  The accuracy of tumor candidate
classification with the random forest approach was 90\%\,(Fig.\,\ref{fig:fig4}).

\section{Conclusions}
\label{sec:conclusions}
We have implemented a fully automatic method for liver tumor
segmentation in CT scans based on three steps: liver segmentation by
classification of all voxels of the CT volume, tumor segmentation by
classification of liver voxels, and tumor candidate filtering by
classification of connected components. Altogether, our method
ranked second according to the volumetric overlap.
We would like to emphasize that the evaluation of the automatic tumor
segmentation depends on the application scenario and that the Dice score
only gives \mbox{limited} information.  The best evaluation criterion may
have to differentiate between a precise tumor volumetry required for
e.g. radioembolisation planning and a tumor detection step that checks
whether tumors are present.

In our future work, we plan to investigate 3D network architectures
and multi-label approaches in order to overcome the problems faced by
our current method.

\bibliographystyle{IEEEbib}
\bibliography{litsen}

\end{document}